\documentclass[10pt,twocolumn,letterpaper]{article}

\usepackage{3dv}
\usepackage{times}
\usepackage{epsfig}
\usepackage{graphicx}
\usepackage{amsmath}
\usepackage{amssymb}

\threedvfinalcopy 

\def\threedvPaperID{****} 

\ifthreedvfinal\fi
\begin{document}

\title{3D Meta-Segmentation Neural Network}

\author{Yu Hao, Yi Fang \\
NYU Multimedia and Visual Computing Lab\\
New York University Abu Dhabi, Abu Dhabi, UAE\\
New York University, New York, USA \\
\texttt{\{yh3252, yfang\}@nyu.edu} \\
}

\maketitle

\begin{abstract}
   Though deep learning methods have shown great success in 3D point cloud part segmentation, they generally rely on a large volume of labeled training data, which makes the model suffer from unsatisfied generalization abilities to unseen classes with limited data. To address this problem, we present a novel meta-learning strategy that regards the 3D shape segmentation function as a task. By training over a number of 3D part segmentation tasks, our method is capable to learn the prior over the respective 3D segmentation function space which leads to an optimal model that is rapidly adapting to new part segmentation tasks. To implement our meta-learning strategy, we propose two novel modules: meta part segmentation learner and part segmentation learner. During the training process, the part segmentation learner is trained to complete a specific part segmentation task in the few-shot scenario. In the meantime, the meta part segmentation learner is trained to capture the prior from multiple similar part segmentation tasks. Based on the learned information of task distribution, our meta part segmentation learner is able to dynamically update the part segmentation learner with optimal parameters which enable our part segmentation learner to rapidly adapt and have great generalization ability on new part segmentation tasks. We demonstrate that our model achieves superior part segmentation performance with the few-shot setting on the widely used dataset: ShapeNet.
\end{abstract}

\section{Introduction}

Point cloud part segmentation is a fundamental computer vision problem, which aims to estimate the part category of each point in the 3D point cloud representation. Though 3D point cloud part segmentation has many real-world applications such as autonomous driving and robotics, it is a challenging task due to the unstructured and unordered characteristics of point clouds. Recently, a number of data-driven deep learning models \cite{qi2017pointnet, qi2017pointnet++, li2018so, qi2016volumetric, wang2017cnn, li2018pointcnn,  su2018splatnet} achieved promising performance and gained popularity in learning 3D point cloud part segmentation tasks compared to traditional approaches. 
\begin{figure*}
\centering
\includegraphics[width=17cm]{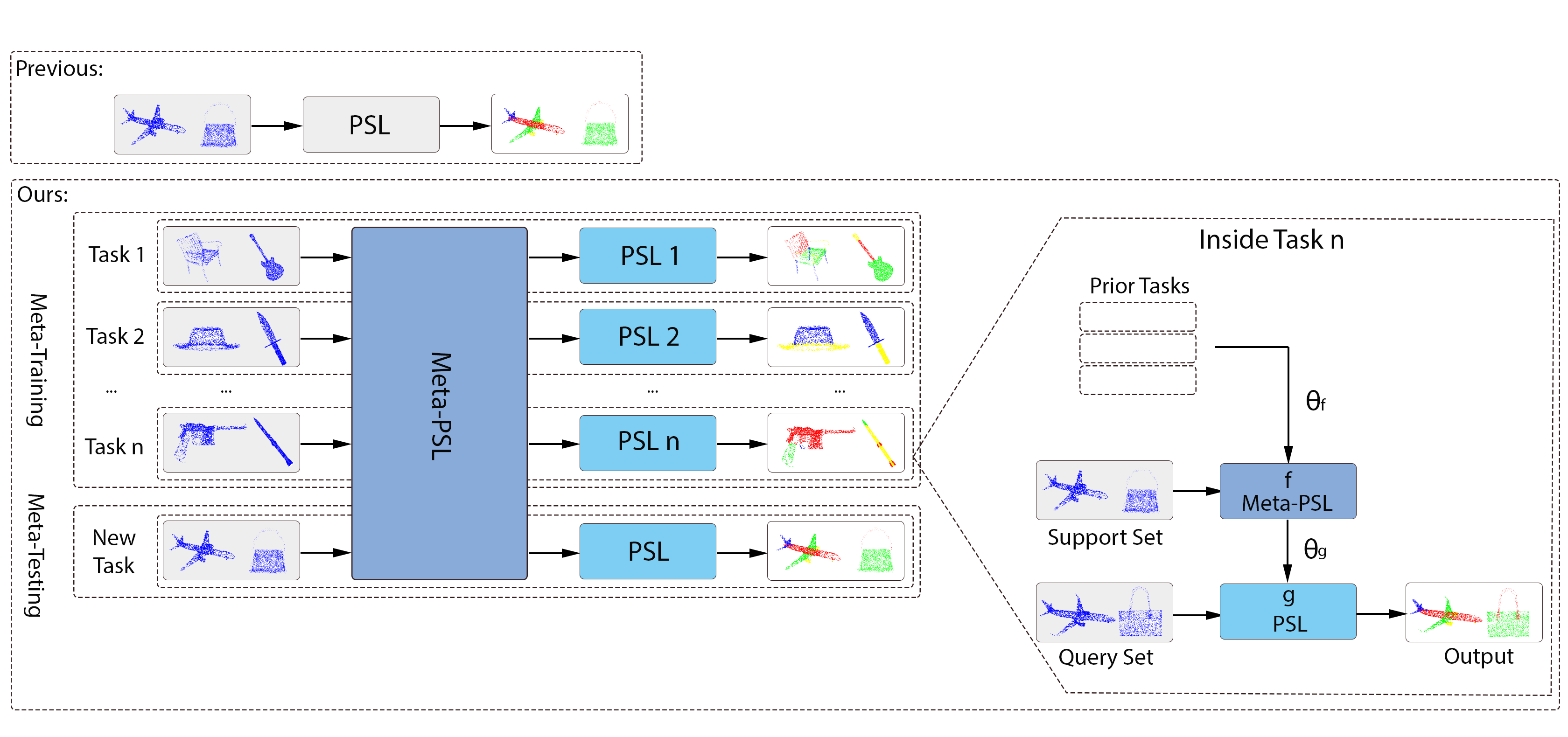}
\caption{Comparison of the training strategy between conventional learning-based methods and our novel meta-learning strategy. Previous learning-based methods regard the part segmentation as a task for the given point clouds. In comparison, by training over a number of 3D part segmentation tasks, our meta-part segmentation learner(Meta-PSL) with $\theta_f$ is able to dynamically update part segmentation learner(PSL) with optimal parameters $\theta_g$ which enable our part segmentation learner to rapidly adapt to new part segmentation tasks.}
\label{fig:1}
\end{figure*}

Qi et al. proposed PointNet \cite{qi2017pointnet} which applies a symmetric function to segment raw point clouds. DGCNN \cite{wang2019dynamic} proposed by Wang et al. focuses on improving the performance by using a module called the EdgeConv that can capture local structures. Such these learning-based methods \cite{li2019cross, li2019pc, wang2017unsupervised, wang2018unsupervised} achieve great success in many fields. However, their success heavily relies on the fact that the optimization process requires iterative training over a large number of labeled data which is expensive and hard to collect. These models perform unsatisfied only given a few examples of novel classes. Moreover, the classifier those approaches use is hard to generalize to new classes due to the small amount of data. To alleviate these limitations, we firstly propose a novel model, named Meta-3DSeg, that is able to capture the information of task distribution and rapidly generalize from limited examples on novel classes.

\begin{figure}
\centering
\includegraphics[width=8.3cm]{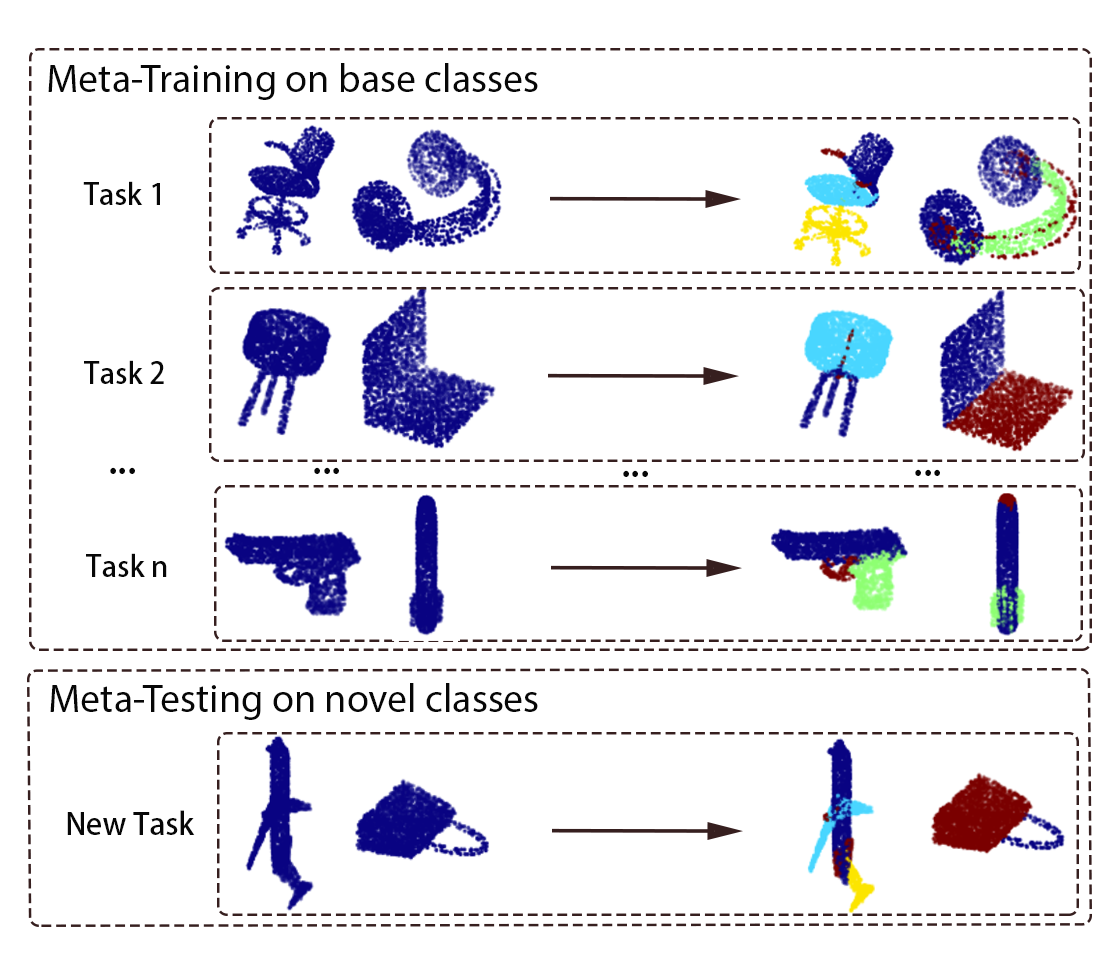}
\caption{Few-shot setting. Our few-shot setting includes two processes: Meta-Training refers to the training and testing process over multiple tasks. Meta-Testing refers to the training and testing process on the new task. Each task is a 2-way-1-shot part segmentation problem, which means each training task contains a support set with only one shape each from 2 different base categories.}
\label{fig:2}
\end{figure}

\begin{figure*}
\centering
\includegraphics[width=17cm]{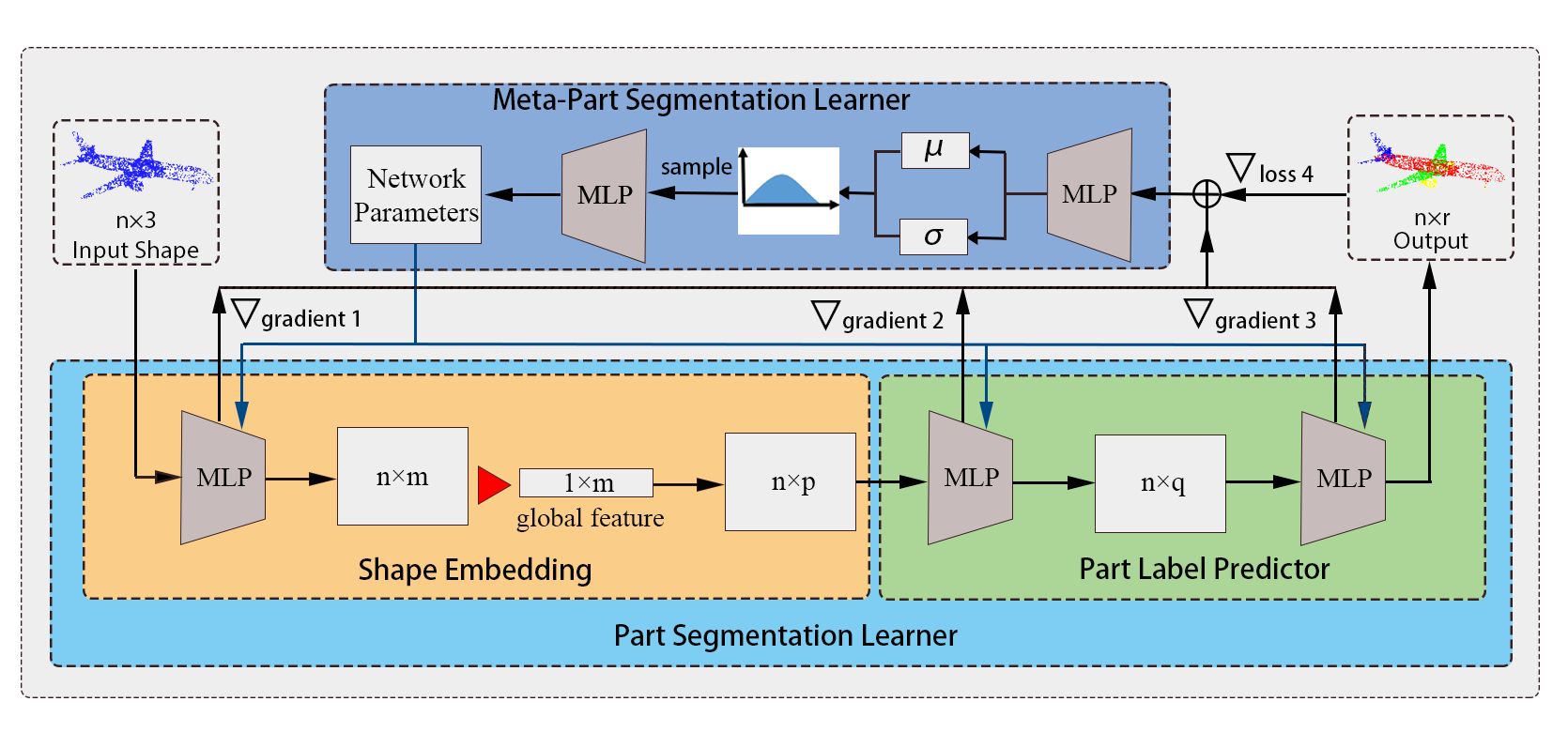}
\caption{Main Pipeline. Our Meta-3DSeg includes two key modules: part segmentation learner and meta-part segmentation learner. The part segmentation learner which consists of learning shape descriptor and part label predictor is trained to complete a specific part segmentation task in the few-shot scenario. The parameters of MLPs in the part segmentation learner are predicted from the meta-part segmentation learner. The meta-part segmentation learner takes the gradient loss and segmentation loss obtained from the part segmentation learner as input. The output of the meta-part segmentation learner is the estimated network parameters of the part segmentation learner.}
\label{fig:3}
\end{figure*}

As shown in Figure \ref{fig:1}, previous learning-based methods regard the 3D shape segmentation function as a single task for the given point clouds. Under this circumstance, the 3D shape segmentation function learner performs well when there are a large number of labeled data for training. However, it is costly and challenging to acquire such numerous high-quality 3D point cloud dataset with part labels. Therefore, in comparison to previous learning-based methods, we proposed to formalize the learning of a 3D shape segmentation function space as a meta-learning problem, aiming to predict a 3D segmentation model that can be quickly adapted to new shapes with limited training data. Instead of learning the distribution over data, our training strategy gains the advantage by training over the distribution of tasks. After training, our meta-part segmentation learner is able to capture the prior knowledge from the distribution of tasks and guide the learning process of the part segmentation learner with optimal parameters. In this way, our part segmentation learner can rapidly adapt to new part segmentation tasks. We describe our few-shot setting for 3D point cloud part segmentation in Figure \ref{fig:2}. Our few-shot setting includes two processes: meta-training and meta-testing. Meta-training refers to the training and testing process over multiple tasks. Meta-testing refers to the training and testing process on the new task. Figure \ref{fig:3} illustrates the main pipeline of our Meta-3DSeg which consists of two components: part segmentation learner and meta-part segmentation learner. Given the input point cloud, the part segmentation learner starts with learning the local and global feature information using a learning shape descriptor. Then a novel part label predictor is proposed to predict the part specific label for each point in the point cloud. As for the second component, the meta-part segmentation learner uses the gradient loss and segmentation loss obtained from the part segmentation learner to learn the pattern of each task and provide optimal parameters for the part segmentation learner. We add a variational auto-encoder(VAE) to learn the task distribution of the input 3D shape, which can be further sampled as the function prior over task distribution for the learner predictor to estimate the optimal weights of the 3D segmentation learner. Our contributions are listed as below:

\begin{itemize}
\item In this paper, we firstly propose a novel meta-learning strategy that differentiates our method from learning-based approaches for training a model to address challenging issues for few-shot 3D point cloud part segmentation in 3D computer vision.

\item In this paper, to implement our novel meta-learning strategy, our Meta-3DSeg consists of two modules: part segmentation learner and meta-part segmentation learner. The part segmentation learner is trained to complete a specific part segmentation task in the few-shot scenario. The meta-part segmentation learner is trained to learn the general information of task distribution.
 
\item In this paper, with our meta-learning strategy, we firstly define the part segmentation estimation of point clouds as a task. Based on the learned information of task distribution, our meta-part segmentation learner is able to dynamically update optimal parameters of the part segmentation learner so that our part segmentation learner can rapidly adapt to new part segmentation tasks.
 
 
\end{itemize}

\section{Related Works}
\subsection{Point Cloud Part Segmentation}

3D point cloud part segmentation is an important 3D computer vision task that aims to predict the part-specific label for each point of a point cloud. As we all know, traditional architectures such as voxels and image grids have regular data format. However, the representation of point clouds is unstructured and unordered which makes the learning of point-wise labels very challenging. Early works typically transform point clouds data to regular voxel grids \cite{qi2016volumetric, wang2017cnn, choy20163d} and multi-view images \cite{su2015multi, qi2016volumetric} so that this task can be solved by using strong supervisions. PointNet \cite{qi2017pointnet} proposed by Qi et al. is the first work that applies a symmetric function to segment raw point clouds. However, such this method overlooks the local information embedded in the neighboring points. To address this problem, PointNet++ \cite{qi2017pointnet++} and SO-Net \cite{li2018so} improve the performance by extending PointNet hierarchically and focusing on capturing the local information. Moreover, Wang et al. proposed DGCNN \cite{wang2019dynamic} which uses a module called the EdgeConv that can capture local structures. Although these approaches achieved promising performance, large amounts of training data are required since they mostly directly predict 3D shape segmentation in the training process. In real-world scenarios, training data is expensive and hard to acquire which limits the use of those strong supervised methods.

\subsection{Meta-learning methods}

Traditional learning-based methods with a small amount of training datasets can severely lead to overfitting and unsatisfied performance. Meta-learning \cite{andrychowicz2016learning, zhou2018deep, santoro2016meta}, also known as learning to learn, has received a lot of attention and achieved substantial progress recently, which aims to learn new skills rapidly and efficiently only given a few samples. Parameters prediction \cite{finn2017model, finn2018probabilistic, lee2018gradient} is one of the strategies in meta-learning, which refers to a network that is trained to update the parameters of another network. The overall network can be more adaptive to new tasks since the first network can encode the information learned on multiple tasks to the second network. Finn et al. proposed MAML \cite{finn2017model} which starts multiple tasks at the same time and learns a common best base model by different tasks. Vinyals et al. proposed the Matching network \cite{vinyals2016matching} which learns weighted nearest neighbor scores from the input images. Prototypical networks \cite{snell2017prototypical} proposed by Snell et al. use an embedding function to map the input data into a metric space so that the similarity of prototype representations of each class can be used to finish the few-shot classification task. Ravi et al. \cite{ravi2016optimization} use the LSTM to learn an update rule for training a neural network by replacing the stochastic gradient descent optimizer in few-shot learning. In 3D computer vision, Littwin et al. \cite{littwin2019deep} achieve satisfying performance in the 3D shape representation task by mapping the input point cloud to the parameters of a deep neural network. In this paper, we first propose a meta-learning strategy which use a meta-part segmentation learner to learn the general information over a variety of part segmentation tasks so that it can estimate the parameters of the part segmentation learner and make the part segmentation learner's network rapidly adapt to new part segmentation tasks. 

\section{Methods}\label{headings}
We introduce our approach in the following sections. In section \ref{set1}, we introduce the problem statement of learning-based part segmentation methods. Section \ref{set2} illustrates the network structure of our part segmentation learner. We explain the meta-part segmentation learner network in section \ref{set3}. The loss function is defined in section \ref{set4}. 

\subsection{Problem Statement}\label{set1}
    We define our meta-learning formulation at first. For a given training dataset $\bold{D}_{train}$, traditional machine learning strategy optimize the parameters $\theta$ and evaluate its generalization ability on the given testing dataset $\bold{D}_{test}$. We define the optimization task of the traditional learning-based methods which directly use unordered point clouds as input at first. Giving a training dataset $\bold{D}=\{P_i\} $, where $ P_i \subset \mathbb{R}^3$, $P_i$ denotes the input point clouds. We aim to obtain a parametric function $g_{\theta}(P_i)$ using a neural network structure which can predict the part label of the specific point. The traditional learning-based methods optimize the $\theta$ as follows:

\begin{equation}
\begin{split}
\bold{\theta^{*}} =\mathop{\arg\min}_{\theta} [\mathbb{E}_{(P_i)\sim \bold{D}}[\mathcal{L}(g_{\theta}(P_i))]
\end{split}
\end{equation}
where $\mathcal{L}$ represents the pre-defined loss function.

However, in our meta-training strategy, there are a number of similar tasks in the meta-training process. We define the training dataset in each task as the support set $\bold{D}_{support}$ and the testing dataset in each task as the query set $\bold{D}_{query}$ which differentiate our meta-learning strategy from traditional learning-based methods. Thus, we focus on the meta training dataset $\bold{M}_{meta-train}=\{\bold{D}_{support_i},  \bold{D}_{query_i}\}$ which consists of multiple regular support and query sets. The meta testing dataset $\bold{M}_{meta-test}=\{\bold{D}_{support_{new}} \\, \bold{D}_{query_{new}}\}$ consists of the new support and query set. In this paper, our part segmentation learner $g$ includes two sets of parameters: $\theta_t$ and $\theta_m$. $\theta_t$ is trained on the support set in the new task, but $\theta_m$ is predicted by another parametric function $f(s_x)$ which is called meta-part segmentation learner. Thus, we have the desired segmentation estimation function $g_{(\theta_t,\theta_m)}(P_i)$ and we have $\theta_m = f(s_x)$, where $s_x$ refers to the weighted gradient loss and segmentation loss generated from the part segmentation learner and we will discuss it in section \ref{set3}. For a given training data set $\bold{D}$, we have:

\begin{equation}
\begin{split}
\bold{\theta_t^{*}, \theta_m^{*}} =\mathop{\arg\min}_{\theta_t, \theta_m}[\mathbb{E}_{(P_i)\sim \bold{D}}[\mathcal{L}(g_{(\theta_t, f(s_x))}(P_i))]
\end{split}
\end{equation}

As for N-way-K-shot task in few-shot setting, each support set $\bold{D}_{support}$ consists of K shapes with N categories. Therefore, $\bold{D}_{support}$ contains $K*N$ shapes in total and $\bold{D}_{query}$ contains all the shapes for evaluation. 
We describe the meta-training process firstly. As for each task in the meta-training process, we train our part segmentation learner on each dataset $\bold{D}_{support}$ in $\bold{M}_{meta-train}$ and evaluate the part segmentation learner on each dataset $\bold{D}_{query}$ in $\bold{M}_{meta-train}$. Specifically, during the training process, we trained our meta-part segmentation learner and part segmentation learner over multiple similar tasks on the training process. Then we fixed our part segmentation learner and trained our meta-part segmentation learner on the testing process. Note that we use the gradient loss and segmentation loss generated from the part segmentation learner as input to train our meta-part segmentation learner. 

 During the meta-testing process, as for the new task, we train our part segmentation learner on the support set $\bold{D}_{support}$ in $\bold{M}_{meta-test}$ and evaluate our part segmentation learner on the query set $\bold{D}_{query}$ in $\bold{M}_{meta-test}$. Note that the network parameters of the part segmentation learner include two components: trained weight and meta-learned weight. Trained weight is optimized on the support set in the new task. Meta-learned weight is predicted by the meta-part segmentation learner which is trained on the meta-training process.

\subsection{Part Segmentation Learner}\label{set2}
The part segmentation learner includes two modules: learning shape descriptor (\ref{s1}) and part label predictor (\ref{s2}). We will discuss these two modules in the following subsections. 

\subsubsection{Shape Embedding}\label{s1}

For the input point clouds, the shape embedding is an MLP-based neural network that can extract shape features and capture the geometric information. Formally, let $P_i $ denotes the input point clouds and $f_x  \subset \mathbb{R}^{m}$  denotes the feature of $x$, $\forall x\in P_i$. We define the encoding network $g_1: \mathbb{R}^{3} \to \mathbb{R}^m$ which uses multi-layer perceptrons (MLP) with ReLu activation function for feature extraction between the input point cloud and local feature, where m is the dimension of output layer. Then we use a max-pooling layer to aggregate the information from the local feature. We have:

\begin{equation}
\begin{split}
f_{x}= g_1(x)_{x \in P_i}
\end{split}
\end{equation}

\begin{equation}
\begin{split}
g_{x}= \text{Maxpool} \{g_1(x)\}_{x \in P_i}
\end{split}
\end{equation}
where Maxpool denotes an element-wise max pooling function. 

The embedding learning shape descriptor is combined by local feature, global feature and point coordinates. Specifically, $\forall x\in P_i$, we concatenate the learned local feature $f_x$ and learned global feature $g_x$ with the coordinates $x$ as the combined feature$[f_x, g_x, x] \in \mathbb{R}^{p}$. Thus,the shape descriptor of input point cloud $P_i$ is: $\{ [f_x, g_x, x]\}_{x\in P_i}$. Note that the weights in $g_1$ includes two component: the trained weights $\theta_t^1$ and the meta-learned weights $\theta_m^1$. The weights for $g_1$ is the element-wise summation of $\theta_t^1$ and $\theta_m^1$. 

\subsubsection{Part Label Predictor} \label{s2}
In this section, we introduce the network structures for part label prediction. In order to learn the point information between local feature, global feature and point coordinates, we define the decoding network $g_2: \mathbb{R}^{(p)} \to \mathbb{R}^{(q)}$, where $p=2m+3$ and q is the dimension of output layer. $g_2$ is a non-linear MLP-based function. The weights for $g_2$ in part label predictor is the element-wise summation of the trained weights $\theta_t^2$ and the meta-learned weights $\theta_m^2$. $\forall x\in P_i$, we denote the point information feature $p_x$ as:

\begin{equation}
\begin{split}
p_{x}= g_2([f_x, g_x, x])_{x \in P_i}
\end{split}
\end{equation}
where [,] denotes the operation of concatenation. 

After we obtained the point feature $\{p_x\}_{x\in P_i}$ from decoding network $g_2$, we define the segmentation network $g_3: \mathbb{R}^{(q)} \to \mathbb{R}^c$ to learn the point feature for each point in the point cloud, where c refers to the number of part categories. $g_3$ is a non-linear MLP-based function. We note the trained weights in $g_3$ as $\theta_t^3$ and the meta-learned weights in $g_3$ as $\theta_m^3$. The weights for $g_3$ is the element-wise summation of $\theta_t^3$ and $\theta_m^3$. We have the predicted part label $o_x$ as:

\begin{equation}
\begin{split}
o_{x}= g_3([p_x])
\end{split}
\end{equation}

Therefore, the weights $\theta_t^1, \theta_t^2, \theta_t^3$ are learned from training process. The weights $\theta_m^1, \theta_m^2, \theta_m^3$ are predicted from the meta-part segmentation learner which we will discuss in next section.

\subsection{Meta-Part Segmentation Learner}\label{set3}

For the learning shape discriptor $[f_x, g_x, x] \in \mathbb{R}^{(p)}$, we use a multi-layer MLP-based function $f_1: \mathbb{R}^p \to \mathbb{R}^k$ to learn the part specific score of each point in the input shape, where k is the dimension of output layer:

\begin{equation}
\begin{split}
\bold{\lambda_{x}}= f_1([f_x, g_x, x])
\end{split}
\end{equation}

Based on the part specific score $\lambda_{x}$ of each point, we use it as the weight to balance among the gradient loss $\nabla_x$ and segmentation loss $l_x$ we obtained from the part segmentation learner. We define the part specific feature $s_{x}$ as:

\begin{equation}
\begin{split}
\bold{s_{x}}= [\lambda_{x}\nabla_x, \lambda_{x}l_x]
\end{split}
\end{equation}

To enable the meta-part segmentation learner to learn the pattern in each task, we introduce a variational auto-encoder (VAE) network. More specifically, we use a multi-layer MLP-based function $f_2: \mathbb{R}^k \to \mathbb{R}^w$ to learn the mean and variance of the task distribution, where w is the dimension of output layer. We denote $(\bold{\mu_{m^i}}, \bold{\sigma_{m^i}})$ as the mean and the standard deviation of task distribution space, where $i=1,2,3$:
\begin{equation}
\begin{split}
\bold{\mu_{m^1}}, \bold{\sigma_{m^1}}, \bold{\mu_{m^2}}, \bold{\sigma_{m^2}},\bold{\mu_{m^3}}, \bold{\sigma_{m^3}}=f_2([s_{x}])
\end{split}
\end{equation}

Therefore, we are able to sample $\theta_m^1 \sim N(\mu_{m_1},\sigma_{m_1})$, $\theta_m^2 \sim N(\mu_{m_2},\sigma_{m_2})$ and $\theta_m^3 \sim N(\mu_{m_3},\sigma_{m_3})$. By doing so, our meta-part segmentation learner is able to learn the task distribution of the input 3D shapes, which can be further sampled as the function prior over task distribution to estimate the optimal weights of the part segmentation learner. Giving a novel part segmentation task, our part segmentation learner can be rapidly adapted by adding the meta-learned weights of $\theta_m^1, \theta_m^2, \theta_m^3$ from the meta-part segmentation learner together with the trained weights of $\theta_t^1, \theta_t^2, \theta_t^3$ which are learned from the support set in the new task.

\subsection{Loss Function}\label{set4}

We define the loss function in this section. For the input point cloud $P_i$, Assuming that we have ground truth part label $\{v_x^*\}_{x\in P_i}$, we use the simple cross entropy loss between predicted part label and ground truth part. We define $v_x$ as the predicted part label. The loss function is defined as:
\begin{equation}
\begin{split}
\mathcal{L}=-\sum_{i=1}^N\{v_x^*(p)log(v_x(p))\}
\end{split}
\end{equation}

\section{Experiments}
In this section, we conduct experiments to demonstrate the effectiveness of our proposed Meta-3DSeg on the ShapeNet dataset. In section \ref{exp1}, we describe the preparation of the dataset in our experiment. Section \ref{exp2} shows the experimental settings. In section \ref{exp3}, we compare different settings and initialization of our model on the ShapeNet part segmentation benchmark to demonstrate the superiority of the Meta-3DSeg. We perform the study on the number of sampled points and shots in section \ref{exp4} and section \ref{exp5}. Further comparison with other learning-based methods is shown in section \ref{exp6}.

\subsection{Dataset preparation} \label{exp1}
We test the performance of our Meta-3DSeg for 3D point cloud part segmentation on the ShapeNet part dataset \cite{yi2016scalable}. The ShapeNet dataset is an open-source collection, which consists of 16,881 shapes from 16 object categories. Each object category has 2 to 5 part labels(50 in total). To ensure a fair comparison, we follow the official split to preprocess our dataset \cite{wu20153d}.

\begin{figure*}
\centering
\includegraphics[width=15cm]{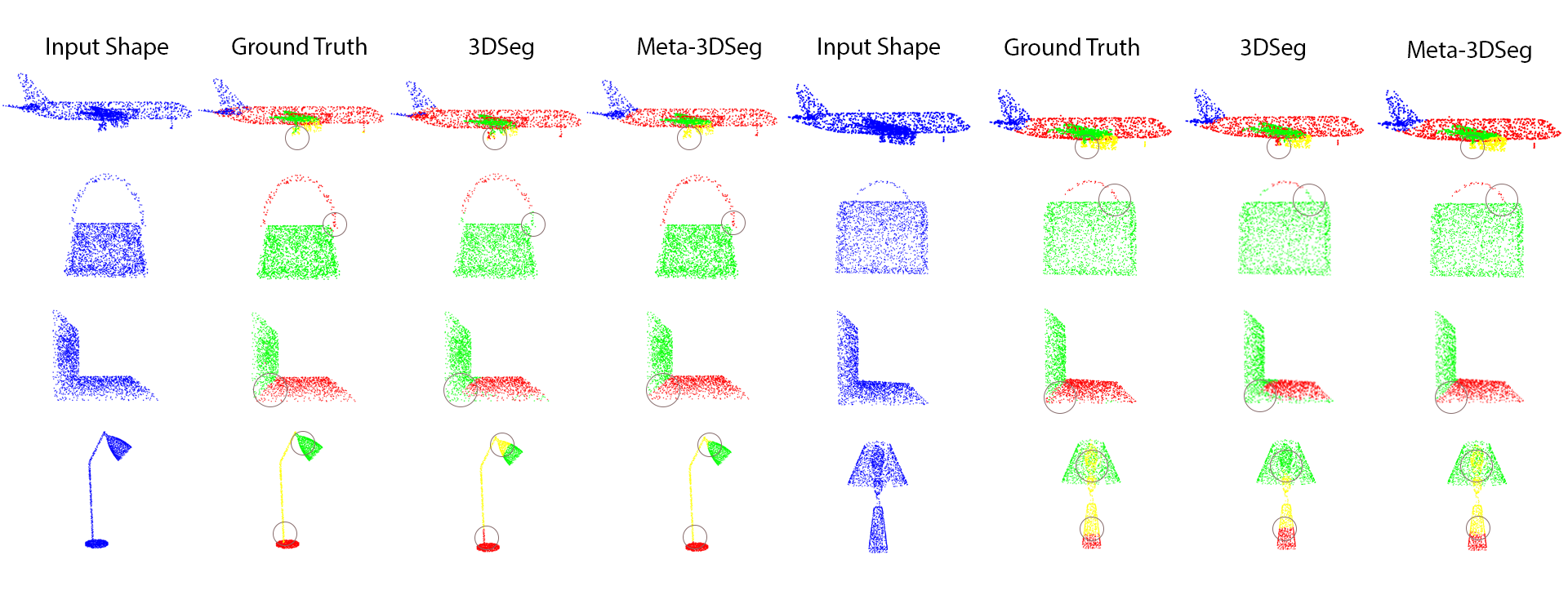}
\caption{ Qualitative results. Comparison of point cloud part segmentation on novel categories of the ShapeNet dataset.}
\label{fig:4}
\end{figure*}

\subsection{Experimental settings} \label{exp2}
In the Meta-3DSeg, the batch size is set to 1, 5, 10 according to the shot we set. We use Adam optimizer as our optimizer. For the part segmentation learner, the learning shape descriptor includes 5 MLPs with dimensions of (64, 128, 128, 512, 2048). The part label predictor network includes 3 MLPs with dimensions of (256,256,128) and one fully connected layers with dimensions of (50) to predict the part label of each point. For the meta-part segmentation learner, we use 2 MLPs with dimensions of (256,128) and one fully connected layers with dimensions of (50) for predicting part attention score and two fully connected layers with dimensions of (10, 1) for estimation of the weights in part segmentation learner. We use the ReLU activation function and implement batch normalization for every MLP layer except the last output layer. We set the learning rate as 0.001.

For evaluation of point cloud part segmentation performance, we use mean IoU (Intersection-over-Union) on points as our metric. Given a particular shape, the average IoU is computed by the ground truth and prediction for each part type. We calculate mIoU for a specific category as the average IoUs of all part types in that category. As for the final mIoU of the category, we compute the average value of mIoU over all shapes in that category.

\begin{table}
  \caption{ Quantitative result. We conduct the ablation study on the ShapeNet dataset.}
  \label{tt1}
  \centering
  \begin{tabular}{lll}
    \hline                  
    Models     & mIoU     & ACC \\
    \hline
    PSL with Weight-Setting-A& 63.3$\%$   & 72.2$\%$     \\
    PSL with Weight-Setting-B & 76.9$\%$  & 86.3$\%$      \\
    PSL with Weight-Setting-C & 77.2$\%$  & 86.9$\%$       \\
    PSL with Weight-Setting-D & 78.8$\%$  & 88.1$\%$      \\
    \hline
  \end{tabular}
\end{table}

\subsection{Ablation study} \label{exp3}
We conduct a series of experiments to verify the effects under different model initialization and settings. 

\paragraph{Experiment setting:} 

For 16,881 shapes from 16 categories in the ShapeNet dataset, we randomly select 10 shapes of the airplane category in the original training dataset as the support set and all shapes of the airplane category in the original testing dataset as the query set for the new task. We sample 2048 points from each point cloud. To show the efficiency of our proposed Meta-3DSeg, we test three settings of our model for comparison. In the weight-setting-A, without our meta-learning strategy, the network parameters of the part segmentation learner are pre-trained using all shapes except for the airplane category in the original training dataset and optimized using support set in the new task. In the weight-setting-B, we directly use MLPs to learn the function prior which is utilized to predict the parameters of the part segmentation Learner. the network parameters of the part segmentation learner are meta-learned by the meta-part segmentation learner and then optimized using the support set of the new task. Note that, as for the dataset in each task during the meta-training process, we randomly select 10 shapes of 1 category except for the airplane category in the original training dataset as the support set and all shapes in that category from the original testing dataset as the query set. As for the dataset in the new task during the meta-testing process, we randomly select 10 shapes from the airplane category in the original training dataset as the support set and all shapes in that category from the original testing dataset as the query set. In the weight-setting-C, with our meta-learning strategy, instead of directly using the MLPs to map the input shape to the function prior, we add a variational auto-encoder(VAE) to learn the task distribution of the input 3D shape, which can be further sampled as the function prior over task distribution to estimate the optimal weights of the part segmentation learner. In the weight-setting-D, we use a part-specific network to predict the part-specific score of a particular point cloud which is used as the weight to balance the gradient loss and segmentation loss obtained from the part segmentation learner. Then the corresponding part-specific gradient loss and segmentation loss is used to train our meta-part segmentation learner.

\paragraph{Results:} 
From the first two rows in Table \ref{tt1}, we notice that the experimental results of the weight-setting-A which simple pre-trained on base classes and then directly optimized using 10 shapes in the airplane category achieves 63.3$\%$ mIoU and 72.2$\%$ segmentation accuracy in the airplane category. Compared with the first setting, the results of the weight-setting-B which use meta-learned weights improved 13.6$\%$ mIoU and 14.1$\%$ segmentation accuracy in the airplane category. This indicates that our meta-learning methods can learn the general information over a number of part segmentation tasks and benefit the part segmentation learner from good meta-learned initialized weights. Moreover, we notice that the experimental results of the weight-setting-C are better than the weight-setting-B, which indicates that the task distribution generated by the VAE in the meta-part segmentation learner can gather the prior over the 3D segmentation function space and can be further sampled to predict the optimal parameters of the part segmentation learner. As indicated in row 3 and 4, the experimental results of the weight-setting-D are better than the weight-setting-C, which indicates that the part specific score generated by the part specific network can be appropriate weight to balance among gradient loss and segmentation loss. In the rest of the experiments, we use this setting as our default model.

\begin{table}
  \caption{ Quantitative result. Comparison of the IoU metric of three categories using shapes with different numbers of sampled points.}
  \centering
  \label{tt2}
  \begin{tabular}{llll}
    \hline                  
    Category     & 512     & 1024 & 2048\\
    \hline
    Airplane& 77.9$\%$   & 78.3$\%$   & 78.8$\%$  \\
    Bag& 70.0$\%$   & 72.2$\%$   & 76.4$\%$  \\
    Laptop& 93.6$\%$   & 94.0$\%$   & 94.3$\%$  \\
    \hline
  \end{tabular}
\end{table}

\subsection{Studies on number of sampled points} \label{exp4}
We conduct the experiments to verify the effects using shapes with different numbers of sampled points. 

\paragraph{Experiment setting:} In this experiment, we conduct the
experiments to verify our model’s performance using various numbers of sampled points of 3D shapes. As for the number of sampled points in 3D shape, we choose three different levels: 512, 1024 and 2048.

\paragraph{Results:} 
As shown in Table \ref{tt2}, when the
number of sampled points in each shape decreases from
2048 to 1024, the IoU on the airplane category drops 0.5$\%$. When the
number of sampled points in each shape decreases from
2048 to 512, the IoU achieves 77.9$\%$ for the Airplane category. As we can see from Table \ref{tt2}, the performance degradation is not obvious when sampling points from 2048 to 1024 and 512 for the airplane and laptop category.

\begin{table*}
  \caption{ Quantitative result. Comparison of the IoU metric of three categories using shapes with different numbers of shots.}
  \label{tt3}
  \centering
  \begin{tabular}{llllllllll}
    \hline                  
    Methods     &\multicolumn{3}{|c|}{Airplane}     & \multicolumn{3}{|c|}{Bag} & \multicolumn{3}{|c}{Laptop}\\
    \hline
         & 1 shot     & 5 shot & 10 shot& 1 shot     & 5 shot & 10 shot& 1 shot     & 5 shot & 10 shot\\
    \hline
    3DSeg & 32.1$\%$   & 60.2$\%$   & 63.3$\%$& 34.7$\%$   & 56.2$\%$   & 64.9$\%$
    & 31.0$\%$   & 86.7$\%$   & 87.3$\%$ \\
    Meta-3DSeg & 51.8$\%$   & 75.1$\%$   & 78.8$\%$
    & 53.8$\%$   & 72.9$\%$   & 76.4$\%$
    & 41.7$\%$   & 93.2$\%$   & 94.3$\%$  \\
    \hline
  \end{tabular}
\end{table*}

\begin{table}
  \caption{ Quantitative result. Comparison of the IoU metric with other learning-based methods using different categories on the ShapeNet dataset.}
  \label{tt4}
  \centering
  \small
  \begin{tabular}{cccccc}
    \hline                  
    Category & PN & PN++ & PC & WPS-Net & Ours
\\
    \hline
    Airplane& 63.3$\%$ & 62.3$\%$ & 65.1$\%$ & 67.3$\%$& \bf78.8$\%$ \\
    Bag& 64.9$\%$ & 67.4$\%$ & 68.2$\%$ & 74.4$\%$& \bf76.4$\%$ \\
    Cap& 75.2$\%$ & 80.0$\%$ & 80.7$\%$ & 86.3$\%$& 81.1$\%$ \\
    Chair& 73.8$\%$ & 61.6$\%$ & 66.1$\%$ & 83.4$\%$& 80.1$\%$ \\
    Lamp& 63.8$\%$ & 57.8$\%$ & 60.2$\%$ & 68.7$\%$& \bf70.5$\%$ \\
    Laptop& 87.3$\%$ & 94.2$\%$ & 93.7$\%$ & 93.8$\%$& \bf94.3$\%$ \\
    Mug& 80.9$\%$ & 83.1$\%$ & 86.0$\%$ & 90.9$\%$& 90.0$\%$ \\
    Table& 72.2$\%$ & 72.2$\%$ & 72.5$\%$ & 74.2$\%$& \bf74.7$\%$ \\
    \hline
    Mean& 72.7$\%$ & 72.3$\%$ & 74.1$\%$ & 79.8$\%$& \bf80.7$\%$ \\
    \hline
  \end{tabular}
\end{table}

\subsection{Studies on number of shots} \label{exp5}
In this section, we conduct the experiments to verify the effects using shapes with different numbers of shots in each category. 

\paragraph{Experiment setting:} In this experiment, we validate our proposed model using various numbers of shots in each category. To prove the effectiveness of our Meta-3DSeg, we mainly use the non-meta and meta-learning setting. In the non-meta (3DSeg) setting, we only use the learning shape descriptor and part label predictor in this case. In the meta-learning setting (Meta-3DSeg), we organically integrate our meta-learning modules to the 3DSeg. We conduct experiments on airplane, bag and laptop categories using 1, 5, 10 shots. 

\paragraph{Results:} 
Table \ref{tt3} shows the quantitative results of the non-meta and meta-learning setting of our model with 1, 5, 10 shots. In comparison with 3DSeg, our Meta-3DSeg achieves higher IoU on the airplane, bag and laptop categories with 1, 5, 10 shots. This indicates that our meta-learning strategy can learn the general information over a number of part segmentation tasks and benefit the part segmentation learner from good meta-learned network parameters. In addition, as illustrated from the qualitative results shown in Figure \ref{fig:4}, our method is more accurate in estimating part-specific labels of point clouds. The part segmentation results in row 4 indicate that there are some points(brown circled region) in the shade area and some points in the base area that are mistakenly labeled as the pole by the 3DSeg. In contrast, with the meta-part segmentation learner, our Meta-3DSeg is able to better align these challenging points to the right part labels.

\subsection{Comparisons with learning-based methods} \label{exp6}
In this section, we conduct the experiments with other learning-based supervised methods. 

\paragraph{Experiment setting:} In this experiment, we evaluate the
overall part segmentation performance of our proposed model under the setting of 1-way-10-shot and yield the performance compared to current state-of-the-art learning-based methods, where PN, PN++, PC, WPS-Net refer to PointNet \cite{qi2017pointnet}, PointNet++ \cite{qi2017pointnet++}, PointConv \cite{wu2019pointconv}, WPS-Net \cite{wang2020few} respectively. For a fair comparison, we use 10 same shapes randomly selected from the dataset as the support set in the new task. Note that we report the performance of these classical learning-based models which are pre-trained with all samples in the base categories for reference since our method is trained using shapes in base categories during the meta-training process. 

\begin{figure}
\centering
\includegraphics[width=8.1cm]{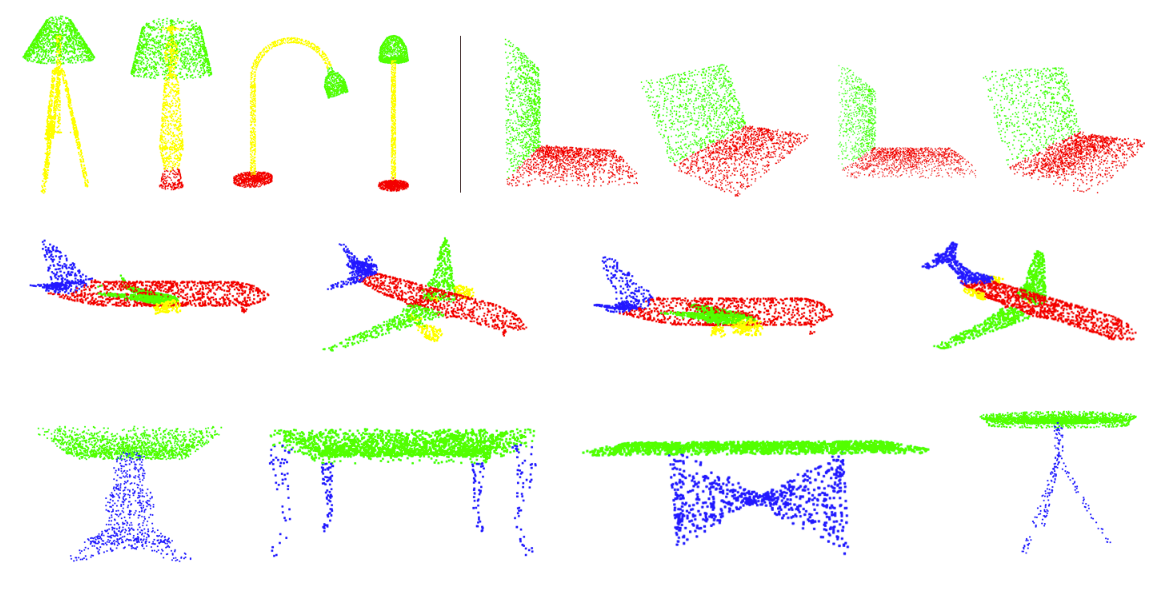}
\caption{ Qualitative results. Randomly selected qualitative results of point clouds part segmentation on several categories of the ShapeNet dataset.}
\label{fig:5}
\end{figure}

\paragraph{Results:} 
As we can see from the Table \ref{tt4}, our Meta-3DSeg achieves higher mIoU (80.7$\%$) compared to WPS-Net (79.8$\%$), PointConv (74.1$\%$), PointNet++ (72.3$\%$) and PointNet (72.7$\%$). As for the airplane, bag, lamp, laptop and table category, our Meta-3DSeg achieves significantly better results with 78.8$\%$ for the airplane category which is better than 67.3$\%$ achieved by WPS-Net. When there are only 10 shapes in each category for training, without our meta-learning strategy, classical learning-based methods are more likely to overfit and behave worse. In this paper, we use the meta-part segmentation learner to learn the general information over multiple similar tasks so that it can provide optimal parameters for the part segmentation learner. Thus, our part segmentation learner is less likely to overfit and have good generalization performance on new part segmentation tasks. Figure \ref{fig:5} shows some randomly selected qualitative results of our Meta-3DSeg on novel categories of the ShapeNet dataset.

\section{Conclusion}
In this paper, we introduce a novel meta-learning strategy to our research community for point cloud part segmentation. Compared with previous learning-based methods, our learning strategy is able to learn the distribution of tasks instead of the distribution of data. With the general information over multiple similar tasks, our Meta-3DSeg is capable to rapidly adapt and have good generalization performance on new tasks. Note that, our proposed meta-learning strategy can hopefully benefit our 3D community by providing an efficient and effective learning algorithm to train the mainstream 3D part segmentation learners (i.e., PointNet). To the best of our knowledge, our method firstly leveraged a novel meta-learning strategy for this task and we experimentally verified the effectiveness of our model and achieved superior few shot part segmentation results on the ShapeNet 3D point cloud part segmentation dataset.


{\small
\bibliographystyle{ieee_fullname}
\bibliography{egbib}
}

\end{document}